\documentclass{article}

\PassOptionsToPackage{numbers,compress}{natbib}
\PassOptionsToPackage{hyphens}{url}
\usepackage[preprint]{neurips_2027_preview}

\usepackage[utf8]{inputenc}
\usepackage[T1]{fontenc}
\usepackage{hyperref}
\usepackage{url}
\usepackage{graphicx}
\usepackage{booktabs}
\usepackage{multirow}
\usepackage{tabularx}
\usepackage{amsmath}
\usepackage{amssymb}
\usepackage{amsfonts}
\usepackage{microtype}
\usepackage{xcolor}
\usepackage{array}
\usepackage{caption}
\usepackage{algorithm}
\usepackage{algorithmic}
\usepackage{flafter}
\usepackage{placeins}

\hypersetup{
  hidelinks,
  pdftitle={Beyond Teacher Likelihood: Group-Calibrated On-Policy Distillation for Long-Context Reasoning},
  pdfauthor={Zhu Zhang; Jixun Wang; Xiaoang Xu; Xiaorong Wang; Zihan Zhou; Zhiyuan Wang; Shuo Wang; Chaojun Xiao; Yuezhi Zhou},
  pdfsubject={NeurIPS 2027 style preview based on the latest released format},
  pdfkeywords={on-policy distillation; long-context reasoning; verifier feedback; knowledge distillation}
}

\newcommand{\method}{GC-OPD}

\newcommand{\GCOPDSingleColumnLayout}{}

\newcommand{\GCOPDValidationFigureWidth}{0.98\linewidth}

\title{Beyond Teacher Likelihood: Group-Calibrated On-Policy Distillation for Long-Context Reasoning}

\author{
  \bfseries
  Zhu Zhang\textsuperscript{1,3}\thanks{Equal contribution. Emails: \nolinkurl{zhuzhang24@mails.tsinghua.edu.cn}, \nolinkurl{wangjixun@bupt.edu.cn}.}
  \quad Jixun Wang\textsuperscript{2,3}\footnotemark[1]
  \quad Xiaoang Xu\textsuperscript{2,3}
  \quad Xiaorong Wang\textsuperscript{3}\\
  \bfseries
  Zihan Zhou\textsuperscript{3}
  \quad Zhiyuan Wang\textsuperscript{1}
  \quad Shuo Wang\textsuperscript{1}
  \quad Chaojun Xiao\textsuperscript{1}\thanks{Corresponding authors. Emails: \nolinkurl{xcj@tsinghua.edu.cn}, \nolinkurl{zhouyz@mail.tsinghua.edu.cn}.}
  \quad Yuezhi Zhou\textsuperscript{1}\footnotemark[2]\\[0.45em]
  {\normalfont\small
  \textsuperscript{1}Tsinghua University
  \quad \textsuperscript{2}Beijing University of Posts and Telecommunications
  \quad \textsuperscript{3}OpenBMB}
}

\begin{document}

\maketitle

\begin{abstract}
On-policy distillation (OPD) trains a student on its own responses using dense token-level guidance from a stronger teacher. In long-context tasks, however, token-level teacher support can favor locally plausible responses that omit evidence distributed across the input or violate global task constraints. Task-specific verifiers, in contrast, evaluate task completion at the response level and may return graded rewards that reflect partial success. We diagnose this mismatch on fixed responses from two representative long-context evidence-aggregation tasks. Across longer input ranges, trajectory-level OPD scores become progressively less aligned with verifier rewards, indicating teacher--verifier disagreement. Motivated by this observation, we introduce Group-Calibrated On-Policy Distillation (\method{}). \method{} separately normalizes verifier rewards and trajectory-level OPD scores within each rollout group and uses their difference as a signed teacher--verifier disagreement residual. Relative-advantage-based credit assignment (RACA) distributes this trajectory-level residual across tokens according to their relative OPD advantages while preserving the original OPD signal. Across five long-context benchmarks, post-training with \method{} raises the five-benchmark averages of the official Qwen3-4B and Qwen3-8B checkpoints from 29.08 to 40.47 and from 35.12 to 44.65, respectively. Vanilla OPD reaches 39.31 and 43.56 under the same setup. Controlled ablations show that the signed residual is more effective than either an additional OPD-derived term or direct group-normalized verifier reward addition, while RACA further improves over uniform token allocation. Together, these results demonstrate that group-relative residual calibration can incorporate verifier outcomes without discarding dense token-level guidance. Code is available at \url{https://github.com/SolereZhang/GC-OPD}.
\end{abstract}

\begin{figure}[t]
\centering
\includegraphics[width=\textwidth]{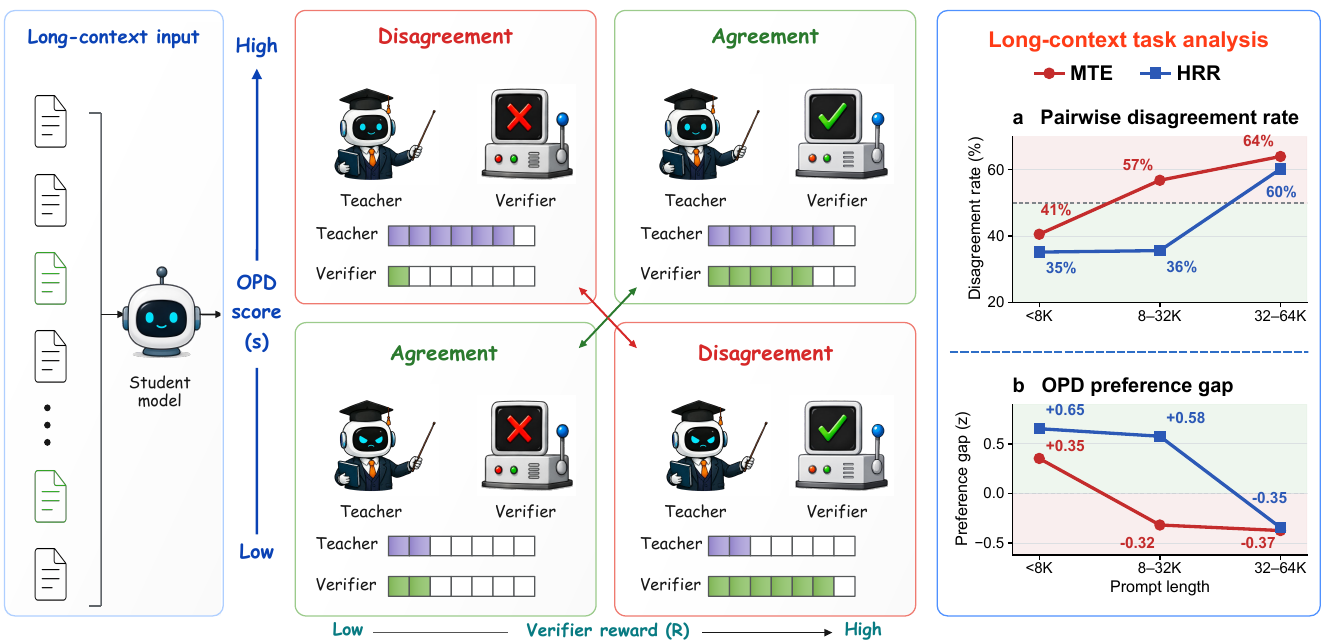}
\caption{Teacher--verifier disagreement in two evidence-aggregation tasks: Multi-Table Extraction (MTE) and High-Recall Retrieval (HRR). Pairwise disagreement rate captures ranking conflicts, while OPD preference gap captures the direction and magnitude of OPD preference along the verifier ordering. Both indicate weaker alignment over longer input ranges. See Section~\ref{sec:obs} for details.}
\label{fig:teacher-verifier-motivation}
\end{figure}

\section{Introduction}

Knowledge distillation transfers knowledge from a strong teacher to a weaker student and is particularly valuable for compact language models~\citep{hinton2015distilling,gu2024minillm,pmlr-v235-ko24c}. On-policy distillation (OPD) applies this paradigm to student-generated responses, providing dense teacher guidance on the student's own generation distribution~\citep{agarwal2024policy}. For each sampled token, OPD derives an advantage from the difference between its teacher and student log-probabilities. This dense signal underlies strong recent OPD methods~\citep{yang2026learning,li2026filter,zhao2026poweropd}, but measures teacher preference rather than verified task success.

This distinction is particularly important for long-context tasks, where correct responses must integrate evidence scattered across distant positions and satisfy global task constraints~\citep{bai2024longbench,hsieh2024ruler,goldman2024really,bai2025longbench}. With longer inputs, student models are more likely to produce locally plausible but globally incomplete responses, as they can struggle to locate and use relevant information when its position or lexical form changes~\citep{liu2024lost,pmlr-v267-modarressi25a}. A response may therefore receive strong teacher support while omitting required evidence. Task-specific verifiers, in contrast, evaluate task completion at the response level and may return graded rewards that reflect partial success~\citep{shao2024deepseekmath}. We call discrepancies between teacher preference and verified task outcomes teacher--verifier disagreement. Figure~\ref{fig:teacher-verifier-motivation} presents diagnostic evidence from a fixed set of generated responses in two representative tasks that share this evidence-aggregation requirement. In both, trajectory-level OPD scores become progressively less aligned with verifier rewards across the analyzed input-length ranges. This trend motivates an interface that preserves dense teacher guidance while correcting response-level preferences inconsistent with verified outcomes.

Existing verifier-aware OPD methods use outcome feedback through objective routing, teacher-context reconstruction, outcome-class calibration, or token-level gating~\citep{zheng2026scope,yu2026multi,hou2026uni,xu2026sg}. These methods establish that verifier feedback complements dense teacher guidance. However, they do not simultaneously preserve graded within-group outcome differences and translate the resulting response-level discrepancy into token-dependent calibration while retaining the original OPD advantage.

To address these limitations, we propose Group-Calibrated On-Policy Distillation (GC-OPD). Following the group-relative comparison in Group Relative Policy Optimization~\citep{shao2024deepseekmath}, GC-OPD separately normalizes verifier rewards and trajectory-level OPD scores within each rollout group. As Figure~\ref{fig:gcopd-overview} shows, their difference forms a signed residual whose direction and magnitude quantify teacher--verifier disagreement. Subtracting the group-normalized OPD assessment focuses calibration on their discrepancy rather than directly adding verifier feedback. Relative-advantage-based credit assignment (RACA) distributes this residual across tokens according to their relative OPD advantages while retaining the original dense OPD signal.

Our contributions are as follows:
\begin{itemize}
    \item We identify a shared pattern across two distributed-evidence long-context tasks: trajectory-level OPD scores become progressively less aligned with verifier rewards over longer inputs, as measured by pairwise disagreement rate and OPD preference gap.
    \item We propose GC-OPD, which retains the original OPD advantage while forming a signed residual between group-normalized verifier rewards and trajectory-level OPD scores. RACA distributes this residual across tokens using relative OPD advantages.
    \item Across two model scales and five long-context benchmarks, GC-OPD achieves the highest average score under the shared setup. Ablations show that the signed residual is more effective than either an additional OPD-derived term or direct group-normalized verifier reward addition, while RACA improves over uniform token allocation.
\end{itemize}

\ifdefined\GCOPDSingleColumnLayout
\begin{figure}[t]
\centering
\includegraphics[width=\textwidth]{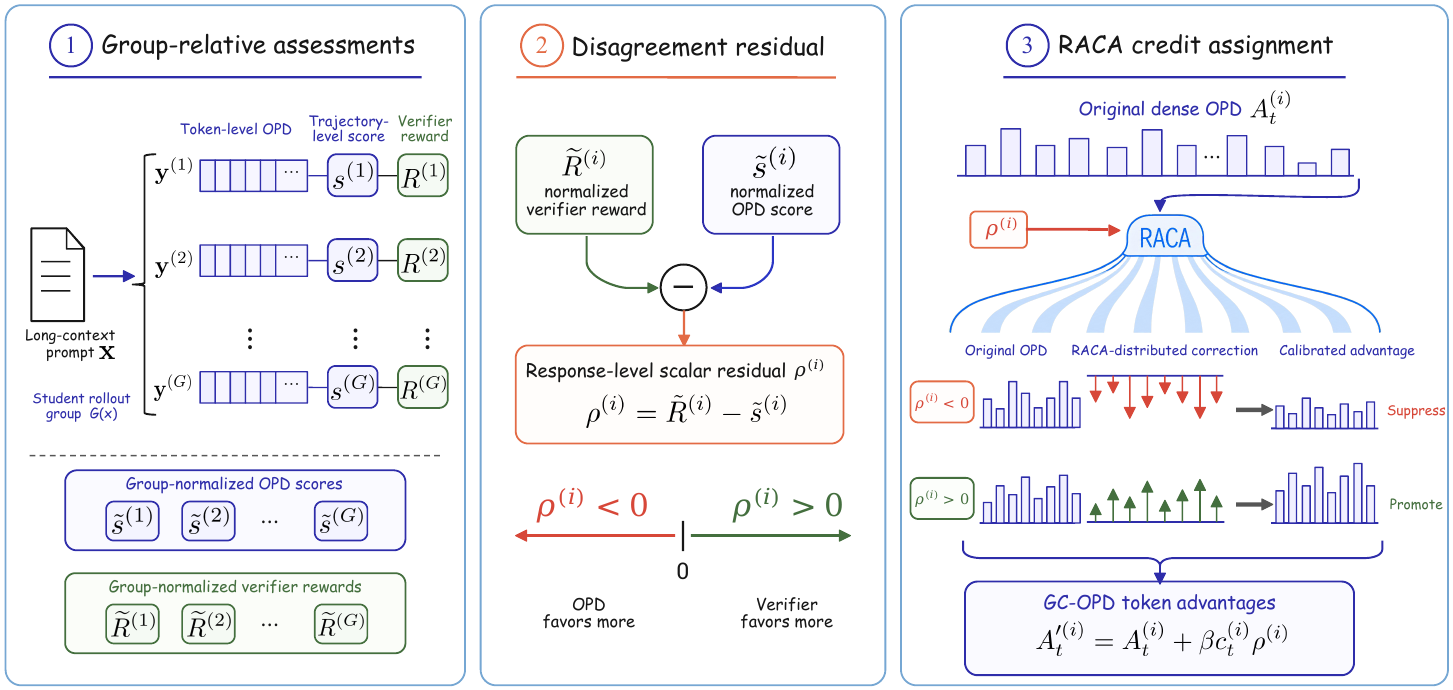}
\caption{Overview of \method{}. GC-OPD computes group-normalized verifier rewards and trajectory-level OPD scores, forms their signed residual, and uses RACA to distribute the residual across response tokens while retaining the original OPD advantage.}
\label{fig:gcopd-overview}
\end{figure}
\else
\begin{figure*}[t]
\centering
\includegraphics[width=\textwidth]{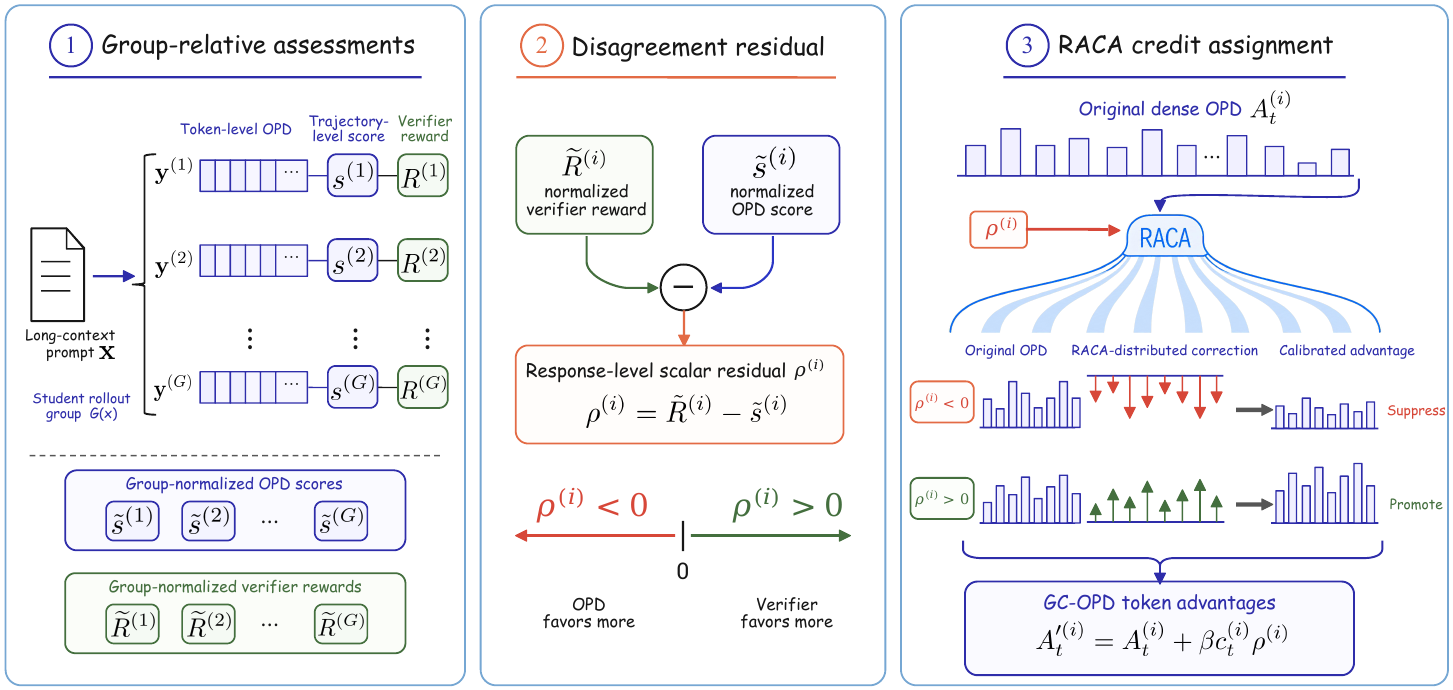}
\caption{Overview of \method{}. GC-OPD computes group-normalized verifier rewards and trajectory-level OPD scores, forms their signed residual, and uses RACA to distribute the residual across response tokens while retaining the original OPD advantage.}
\label{fig:gcopd-overview}
\end{figure*}
\fi

\section{Related Work}

\textbf{Long-context reasoning and verifiable post-training.}
Long-context benchmarks show that nominal context capacity does not guarantee reliable retrieval, integration, or reasoning over evidence distributed across long inputs \citep{bai2024longbench,hsieh2024ruler,liu2024lost,goldman2024really,pmlr-v267-modarressi25a}. Corresponding post-training work improves these capabilities through long-instruction data, length-aware optimization, reinforcement-learning curricula, and task-specific feedback \citep{bai2024longalign,wan2025qwenlong,wang2026loongrl,chen2026longrlvr,zhang2025longreward}. These studies improve the construction of long-context data, curricula, and rewards. Our setting instead begins with an existing task verifier and asks how its response-level assessment should interact with dense token-level OPD guidance.

\textbf{Teacher-centric on-policy distillation.}
MiniLLM and GKD distill on student-generated sequences, exposing the teacher to the student's own generation distribution \citep{gu2024minillm,agarwal2024policy}. ExOPD extrapolates along a teacher--reference direction \citep{yang2026learning}, FiRe-OPD filters trajectories and reweights informative tokens \citep{li2026filter}, and PowerOPD bounds sampled-token rewards for stability \citep{zhao2026poweropd}. Complementary analyses examine teacher--student compatibility and whether large token-level disagreement is learnable, while token-selective distillation concentrates supervision on selected positions \citep{li2026rethinking,wang2026not,kim2026explain}. This line improves how teacher-derived supervision is formed, selected, or optimized. It does not, however, explicitly compare the resulting trajectory-level OPD assessment with a task-specific verifier reward.

\textbf{Verifier-aware on-policy distillation.}
Recent methods connect verifier outcomes to teacher supervision through several interfaces. SCOPE routes correct responses to weighted maximum likelihood and incorrect responses to teacher-perplexity-weighted OPD \citep{zheng2026scope}. MOPD conditions teacher supervision on successful and failed peer responses \citep{yu2026multi}. Uni-OPD calibrates trajectory returns using an outcome-class margin \citep{hou2026uni}. SG-OPD gates token updates by sign consistency and supplements early training with verifier-endorsed teacher rollouts \citep{xu2026sg}. Reward-Weighted OPD selects verifier-passable rollouts and weights their dense teacher supervision \citep{zou2026reward}. These methods introduce verifier feedback through objective routing, teacher conditioning, return calibration, token gating, or trajectory weighting. GC-OPD instead leaves the teacher pass unchanged and forms a response-specific residual from group-relative verifier and OPD assessments. It retains graded within-group differences and the original token-level OPD advantage.

\textbf{Trajectory-level feedback and token-level credit.}
A terminal verifier supplies one response-level reward without specifying its tokenwise influence. Process-supervision methods use human or automatically constructed step labels \citep{lightman2024let,wang2024math}, while VinePPO estimates action advantages from auxiliary continuations \citep{pmlr-v267-kazemnejad25a}. These approaches obtain finer-grained evidence through additional labels or sampling. GC-OPD requires neither step labels nor auxiliary continuations. RACA instead allocates the trajectory-level residual with bounded weights derived from relative OPD advantages without inferring token correctness.

\section{Background and Motivation}

\subsection{On-Policy Distillation}

Let $\mathbf{x}$ denote a long-context input and $\pi_\theta$ the trainable student policy. At the beginning of each update, $\pi_{\theta_{\mathrm{old}}}$ denotes a frozen copy of $\pi_\theta$ used to generate the current rollouts. For each input, we sample a rollout group of $G$ responses, $\mathcal{G}(\mathbf{x})=\{\mathbf{y}^{(i)}\}_{i=1}^{G}$, from $\pi_{\theta_{\mathrm{old}}}$, where $\mathbf{y}^{(i)}=(y^{(i)}_1,\ldots,y^{(i)}_{T^{(i)}})$ has length $T^{(i)}$.

Let $\pi_T$ denote the teacher policy. For each response sampled from $\pi_{\theta_{\mathrm{old}}}$, vanilla OPD defines the token-level advantage of $y_t^{(i)}$ as
\begin{equation}
\begin{aligned}
A^{(i)}_t={}&
\log \pi_T(y^{(i)}_t \mid \mathbf{x},\mathbf{y}^{(i)}_{<t})\\
&-\log \pi_{\theta_{\mathrm{old}}}(y^{(i)}_t \mid \mathbf{x},\mathbf{y}^{(i)}_{<t}).
\end{aligned}
\label{eq:opd-token-advantage}
\end{equation}

Let $h_t^{(i)}=(\mathbf{x},\mathbf{y}^{(i)}_{<t})$ denote the input and response prefix at position $t$. Taking the expectation over a token sampled from the frozen rollout policy gives
\begin{equation}
\mathbb{E}_{y_t^{(i)}\sim\pi_{\theta_{\mathrm{old}}}(\cdot\mid h_t^{(i)})}
\!\left[A_t^{(i)}\right]
=
-\mathrm{KL}\!\left(
\pi_{\theta_{\mathrm{old}}}(\cdot\mid h_t^{(i)})\Vert\pi_T(\cdot\mid h_t^{(i)})
\right).
\label{eq:opd-reverse-kl}
\end{equation}
Thus, $A_t^{(i)}$ is a sampled-token contribution whose expectation equals the negative reverse KL~\citep{gu2024minillm,zhao2026poweropd}. A positive value indicates greater teacher than student support for the sampled token, whereas a negative value indicates the converse. The signal is dense but reflects teacher preference rather than verified task outcome.

During actor optimization, $A_t^{(i)}$ is treated as fixed and weights the clipped token-level objective for $\pi_\theta$, whose importance ratio is computed relative to $\pi_{\theta_{\mathrm{old}}}$. Positive advantages increase the probability of sampled tokens, whereas negative advantages decrease it.

\subsection{Teacher--Verifier Disagreement in Long-Context OPD}
\label{sec:obs}

A task-specific verifier assigns each response a reward $R^{(i)}$, which may encode binary correctness or graded partial success. To compare this response-level outcome feedback with token-level OPD guidance, we aggregate the token-level advantages into a trajectory-level OPD score:
\begin{equation}
s^{(i)}=\frac{1}{T^{(i)}}\sum_{t=1}^{T^{(i)}}A^{(i)}_t.
\label{eq:trajectory-opd-score}
\end{equation}
Averaging avoids direct scaling with response length and summarizes the teacher's mean support for the response relative to the frozen rollout policy.

The trajectory-level OPD score and verifier reward assess different properties of the same response. The former summarizes teacher support relative to the frozen rollout policy, whereas the latter measures verified task outcome. Their orderings within a rollout group can therefore disagree: a response with a higher OPD score may receive a lower verifier reward than another response. We refer to such discrepancies as \emph{teacher--verifier disagreement}. This disagreement is particularly consequential in long-context OPD, where task success can depend on aggregating evidence distributed across distant positions. In this setting, token-level teacher support may favor a locally plausible response even when it omits evidence required by the verifier. Because the two signals also differ in numerical scale, we characterize their agreement through relative comparisons within each rollout group rather than their raw values.

We examine how teacher--verifier disagreement varies with prompt length using a fixed set of generated responses from Multi-Table Extraction and High-Recall Retrieval in GoLongRL~\citep{lv2026golongrl}. Both tasks require aggregating evidence distributed across the input and use graded verifier rewards. The analysis covers 751 Multi-Table prompts and 2{,}908 High-Recall prompts across three prompt-length ranges below 64K tokens. We use Qwen3-8B as the student and Qwen3-30B-A3B-Thinking-2507 as the teacher. We extend the student's context window with YaRN using a scaling factor of 4~\citep{peng2024yarn}, while the teacher uses its native context window. For each prompt, the student generates eight responses, the teacher supplies the token-level log-probabilities used to compute $s^{(i)}$, and the task-specific verifier assigns $R^{(i)}$.

For a prompt $\mathbf{x}$, let $\mathcal{P}(\mathbf{x})=\{(i,j):R^{(i)}>R^{(j)}\}$ contain all response pairs with distinct verifier rewards, oriented toward the higher-reward response. We normalize the trajectory-level OPD scores to zero mean and unit variance within the rollout group and denote them by $\tilde{s}^{(i)}$. Let $\mathbb{I}(\cdot)$ denote the indicator function. The pairwise disagreement rate is
\begin{equation}
d_{\mathrm{pair}}(\mathbf{x})=
\frac{1}{|\mathcal{P}(\mathbf{x})|}
\sum_{(i,j)\in\mathcal{P}(\mathbf{x})}
\mathbb{I}(\tilde{s}^{(i)}<\tilde{s}^{(j)}),
\label{eq:pairwise-disagreement}
\end{equation}
and the OPD preference gap is
\begin{equation}
g_{\mathrm{OPD}}(\mathbf{x})=
\frac{1}{|\mathcal{P}(\mathbf{x})|}
\sum_{(i,j)\in\mathcal{P}(\mathbf{x})}
(\tilde{s}^{(i)}-\tilde{s}^{(j)}).
\label{eq:opd-preference-gap}
\end{equation}
The first metric measures how often the OPD ordering disagrees with the verifier ordering. The second captures the direction and magnitude of OPD preference along the verifier ordering, measured in units of the within-group OPD standard deviation. Positive gaps indicate agreement with the verifier ordering, whereas negative gaps indicate greater OPD support for lower-reward responses. We macro-average both metrics over prompts whose eight responses contain at least two distinct verifier rewards.

Figure~\ref{fig:teacher-verifier-motivation} shows that the pairwise disagreement rate increases over longer prompt-length ranges in both tasks, while the OPD preference gap declines from positive to negative. For Multi-Table Extraction, the disagreement rate rises from 40.6\% below 8K to 64.0\% at 32--64K, and the preference gap declines from $+0.35$ to $-0.37$. For High-Recall Retrieval, the corresponding values change from 35.2\% to 60.2\% and from $+0.65$ to $-0.35$. Within these two tasks, the consistent trends indicate that longer inputs can amplify teacher--verifier disagreement, motivating verifier-based calibration of dense OPD guidance.

\FloatBarrier
\section{Method}

\subsection{Overview}

Group-Calibrated On-Policy Distillation (GC-OPD) calibrates vanilla OPD with response-level verifier feedback while retaining its dense token-level guidance. As illustrated in Figure~\ref{fig:gcopd-overview}, GC-OPD first normalizes verifier rewards and trajectory-level OPD scores within each rollout group, then represents their difference as a signed teacher--verifier disagreement residual. Relative-advantage-based credit assignment (RACA) distributes this trajectory-level residual across tokens according to their relative OPD advantages, and the resulting correction is added to the original token-level OPD advantage.

\subsection{Group-Relative Assessments}

For each response $\mathbf{y}^{(i)}\in\mathcal{G}(\mathbf{x})$, vanilla OPD supplies a token-level OPD advantage $A^{(i)}_t$ for every response token. Equation~\ref{eq:trajectory-opd-score} aggregates these advantages into the trajectory-level OPD score $s^{(i)}$, while the verifier assigns the response a reward $R^{(i)}$. These quantities capture different properties of the response. The trajectory-level OPD score summarizes the teacher's average support relative to the frozen rollout policy, whereas the verifier reward measures verified task outcome.

Since the two signals differ in their raw scales, GC-OPD normalizes them separately within each rollout group. Let $z(\cdot)$ denote within-group z-score normalization, which subtracts the within-group mean and divides by the within-group standard deviation. We define the group-normalized verifier reward and group-normalized trajectory-level OPD score as
\begin{equation}
\begin{split}
\tilde{R}^{(i)} &=z\left (R^{(i)} \right ),\\
\tilde{s}^{(i)} &=z\left (s^{(i)} \right ).
\end{split}
\label{eq:group-relative-signals}
\end{equation}

Normalizing the two signals separately within each rollout group avoids directly comparing their raw values. For graded verifier rewards, $\tilde{R}^{(i)}$ retains the relative spacing of outcomes within the group. Binary verifier rewards, by contrast, indicate only whether each response succeeds or fails. GC-OPD operates on these group-normalized values rather than their induced rankings alone, allowing the magnitude of graded outcome differences to inform the residual.

\subsection{Teacher--Verifier Disagreement Residual}

Given the group-normalized quantities in Equation~\ref{eq:group-relative-signals}, we define the signed residual as
\begin{equation}
\rho^{(i)}=\tilde{R}^{(i)}-\tilde{s}^{(i)}.
\label{eq:residual}
\end{equation}

The sign of $\rho^{(i)}$ indicates the direction of teacher--verifier disagreement. A positive value means that the group-normalized verifier reward for response $\mathbf{y}^{(i)}$ exceeds its group-normalized OPD score, whereas a negative value indicates the opposite. Its magnitude reflects the difference between the two group-normalized assessments.

GC-OPD uses this difference instead of directly adding the group-normalized verifier reward. Direct reward addition can reinforce a relative preference already expressed by OPD. In contrast, the residual vanishes when the two group-normalized assessments coincide, while its magnitude grows with their discrepancy. The residual therefore focuses calibration on differences between the verifier and OPD assessments.

The residual also has the desired ordering under a strict disagreement between the verifier and OPD orderings. For two responses $i$ and $j$ in the same rollout group, let $\sigma_R$ and $\sigma_s$ denote the positive within-group standard deviations. If $R^{(i)}>R^{(j)}$ but $s^{(i)}<s^{(j)}$, then
\begin{equation}
\rho^{(i)}-\rho^{(j)}
=
\frac{R^{(i)}-R^{(j)}}{\sigma_R}
-
\frac{s^{(i)}-s^{(j)}}{\sigma_s}
>0.
\label{eq:pairwise-residual-direction}
\end{equation}

Thus, when the verifier prefers response $i$ but OPD prefers response $j$, the residual assigns a larger value to response $i$. Conversely, reversing the two preferences reverses the residual ordering. This verifier-consistent ordering directs the subsequent token-level correction toward the response with the higher verifier reward.

If either signal has a near-zero within-group standard deviation, the corresponding normalization is not well-defined. We therefore set $\rho^{(i)}=0$ for all responses in that group, reducing GC-OPD to vanilla OPD for the group.

\subsection{Relative-Advantage-based Credit Assignment}

The trajectory-level residual specifies the direction and magnitude of calibration for each response, but not how the correction should be distributed across tokens. RACA uses the token-level OPD advantages within each response as the allocation signal. We first normalize each token's OPD advantage relative to the response mean:
\begin{equation}
u^{(i)}_t=
\frac{A^{(i)}_t-s^{(i)}}
{\sqrt{\frac{1}{T^{(i)}}\sum_{v=1}^{T^{(i)}}
\left(A^{(i)}_v-s^{(i)}\right)^2+\epsilon}},
\end{equation}
where $\epsilon$ is a small positive constant for numerical stability. The resulting relative OPD advantage $u^{(i)}_t$ is positive for tokens whose OPD advantages exceed the response mean and negative for those below it. This normalization preserves the within-response ordering of the signed OPD advantages. Here, $u^{(i)}_t$ serves only as an allocation signal and should not be interpreted as token correctness or causal importance.

RACA maps the relative OPD advantage to a positive, bounded token credit:
\begin{equation}
c^{(i)}_t=1+\tanh\left(\frac{u^{(i)}_t}{2}\right).
\label{eq:raca}
\end{equation}

This mapping is monotonic and satisfies $c^{(i)}_t\in(0,2)$. Tokens with larger relative OPD advantages therefore receive larger credits, while $u^{(i)}_t=0$ yields unit credit $c^{(i)}_t=1$. Because the credit is always positive, it scales but does not reverse the sign of the trajectory-level residual.

The resulting GC-OPD advantage is
\begin{equation}
A'^{(i)}_t=
A^{(i)}_t+\beta c^{(i)}_t\rho^{(i)},
\label{eq:gc-opd-advantage}
\end{equation}
where $\beta\geq0$ is the residual coefficient that controls the strength of calibration. The original token-level OPD advantage remains the base term, and $\beta c^{(i)}_t\rho^{(i)}$ is the token-level residual correction. GC-OPD substitutes $A'^{(i)}_t$ for $A^{(i)}_t$ in the clipped token-level policy objective described in Section~3.1, changing the advantage construction without modifying the actor objective. Setting $\beta=0$ recovers vanilla OPD. Given the OPD advantages and verifier rewards already produced by the training pipeline, GC-OPD adds only group-level aggregation, normalization, and elementwise token transformations. It requires no additional teacher or student forward pass.

\section{Experiments}

We organize our experiments around three questions. First, does \method{} improve upon vanilla OPD under the same long-context training and evaluation setup? Second, does residualizing group-normalized verifier feedback improve upon adding that feedback directly? Third, does relative-advantage-based credit assignment (RACA) improve upon uniform residual allocation?

\subsection{Experimental Setup}

\paragraph{Models.}

We use Qwen3-4B\footnote{\url{https://huggingface.co/Qwen/Qwen3-4B}} and Qwen3-8B\footnote{\url{https://huggingface.co/Qwen/Qwen3-8B}} as student models and Qwen3-30B-A3B-Thinking-2507\footnote{\url{https://huggingface.co/Qwen/Qwen3-30B-A3B-Thinking-2507}} as the teacher model. Both students operate in no-thinking mode. The teacher provides token-level log probabilities for OPD training.

\paragraph{Training data.}

We construct the training set from GoLongRL~\citep{lv2026golongrl} by retaining prompts no longer than 32K tokens. The resulting subset contains 9{,}527 prompts across nine task families. Table~\ref{tab:training-rewards} summarizes their distribution and native verifier metrics. Three task families use binary rewards, while the remaining six use graded rewards. The two largest families, precise long-range retrieval and evidence-grounded reasoning, account for 82.9\% of the training set.

\begin{table}[ht]
\captionsetup{justification=raggedright,singlelinecheck=false,skip=6pt}
\caption{Composition and reward interfaces of the 9{,}527-prompt GoLongRL subset after the 32K filter. ``B'' denotes a binary reward in $\{0,1\}$; ``G'' denotes a graded score in $[0,1]$.}
\label{tab:training-rewards}
\centering
\small
\setlength{\tabcolsep}{3.2pt}
\renewcommand{\arraystretch}{1.05}
\begin{tabular}{@{}l c l@{}}
\toprule
\textbf{Task Family} & \textbf{\# Samples} & \textbf{Native Verifier} \\
\midrule
Precise Long-Range Retrieval & 4,693 & Exact Match (B) \\
Evidence-Grounded Reasoning & 3,204 & Accuracy (B) \\
Multi-Table Extraction & 729 & IoU (G) \\
High-Recall Retrieval & 540 & Set F1 (G) \\
Fragment Matching and Induction & 148 & Subset EM (G) \\
Sequence Reconstruction & 69 & Pairwise Acc. (G) \\
Numerical Reasoning & 54 & Math Verify (B) \\
Graded Retrieval and Ranking & 45 & NDCG (G) \\
Long-Document Summarization & 45 & ROUGE-L (G) \\
\bottomrule
\end{tabular}
\end{table}

\ifdefined\GCOPDSingleColumnLayout\else
\ifdefined\GCOPDSingleColumnLayout
\begin{table}[t]
\else
\begin{table*}[t]
\fi
\caption{Qwen3-4B and Qwen3-8B results on five long-context benchmarks (\%). ``Avg.'' is the unweighted five-task mean. $\dagger$ denotes an implementation of the method's principal training-signal mechanism under the shared setup.}
\label{tab:main-results}
\centering
\small
\setlength{\tabcolsep}{4.0pt}
\begin{tabular}{l l c c c c c c }
\toprule
\multirow{2}{*}{\bf Model} & \multirow{2}{*}{\bf Method} & \multicolumn{6}{c}{\bf Benchmark} \\
\cmidrule(lr){3-8}
& & \bf Avg. & \bf DocMath & \bf Frames & \bf MRCR & \bf CorpusQA & \bf LBv1QA \\
\cmidrule(lr){1-2} \cmidrule(lr){3-8}
\multirow{7}{*}{Qwen3-4B}
& Raw & 29.08 & 43.63 & 26.58 & 22.02 & 6.99 & 46.20 \\
& OPD & 39.31 & 49.37 & \underline{30.95} & 26.90 & 32.22 & \textbf{57.10} \\
& ExOPD$^{\dagger}$ & 38.22 & 47.75 & 29.61 & 25.95 & 31.91 & 55.90 \\
& Uni-OPD$^{\dagger}$ & 38.53 & \textbf{51.88} & 29.98 & 27.93 & 27.36 & 55.50 \\
& PowerOPD$^{\dagger}$ & 38.88 & 49.63 & 30.10 & \textbf{28.36} & \underline{32.52} & 53.80 \\
& FiRe-OPD$^{\dagger}$ & \underline{39.50} & 49.37 & \textbf{31.31} & \underline{28.20} & \underline{32.52} & \underline{56.10} \\
\cmidrule(lr){2-2} \cmidrule(lr){3-8}
& \textbf{\method{} (ours)} & \textbf{40.47} & \underline{50.38} & 30.34 & 27.82 & \textbf{37.99} & 55.80 \\
\cmidrule(lr){1-2} \cmidrule(lr){3-8}
\multirow{7}{*}{Qwen3-8B}
& Raw & 35.12 & 45.88 & 30.95 & 23.87 & 22.19 & 52.70 \\
& OPD & 43.56 & \underline{55.13} & \textbf{34.59} & 30.44 & 39.82 & 57.80 \\
& PowerOPD$^{\dagger}$ & 41.53 & 51.38 & 32.65 & \underline{31.65} & 37.39 & 54.60 \\
& Uni-OPD$^{\dagger}$ & 43.41 & 54.13 & \textbf{34.59} & 25.25 & \underline{42.86} & \underline{60.20} \\
& ExOPD$^{\dagger}$ & 43.49 & 53.12 & 33.50 & \textbf{32.73} & 37.69 & \textbf{60.40} \\
& FiRe-OPD$^{\dagger}$ & \underline{44.01} & 54.00 & \underline{34.34} & 30.96 & 41.64 & 59.10 \\
\cmidrule(lr){2-2} \cmidrule(lr){3-8}
& \textbf{\method{} (ours)} & \textbf{44.65} & \textbf{55.50} & \textbf{34.59} & 31.10 & \textbf{43.77} & 58.30 \\
\bottomrule
\end{tabular}
\ifdefined\GCOPDSingleColumnLayout
\end{table}
\else
\end{table*}
\fi

\fi

\paragraph{Training configuration.}

All post-training methods use the same 100-step optimization budget. At each step, we sample a batch of 32 prompts and generate eight student responses per prompt. The maximum prompt length is 32{,}768 tokens, and response generation is capped at 10{,}240 tokens. Complete optimization and implementation details are provided in the supplementary material.

\paragraph{Evaluation.}

We evaluate DocMath \citep{zhao2024docmath}, Frames \citep{krishna2025fact}, MRCR \citep{vodrahalli2024michelangelo}, CorpusQA \citep{lu2026corpusqa}, and LBv1QA \citep{bai2024longbench}, covering numerical reasoning, multi-hop synthesis, multi-round co-reference, corpus aggregation, and long-context question answering, respectively. Following QwenLong-L1 \citep{wan2025qwenlong}, each student receives at most 120{,}000 input tokens and generates at most 8{,}192 tokens within a 131{,}072-token serving context. We extend the context window with YaRN \citep{peng2024yarn} using a scaling factor of 4. All students remain in no-thinking mode and use the same evaluation prompts and decoding configuration. We report the score on each benchmark and the unweighted mean across the five benchmarks.

\paragraph{Baselines and comparison settings.}

Raw denotes the corresponding officially released Qwen3 checkpoint evaluated before any post-training in our pipeline \citep{yang2025qwen3}. We compare \method{} with Raw, vanilla OPD \citep{agarwal2024policy}, ExOPD \citep{yang2026learning}, Uni-OPD \citep{hou2026uni}, FiRe-OPD \citep{li2026filter}, and PowerOPD \citep{zhao2026poweropd}. Vanilla OPD serves as the dense-distillation control. ExOPD introduces teacher--reference extrapolation, Uni-OPD performs outcome-conditioned margin calibration, FiRe-OPD filters trajectories and reweights token-level signals, and PowerOPD bounds token-level OPD rewards for stability. Daggered rows in Table~\ref{tab:main-results} implement each method's principal training-signal mechanism under our shared long-context setup. To isolate these mechanisms, the runs use the same student initialization, teacher, training data, rollout configuration, 100-step training budget, and evaluation pipeline as \method{}.

\begin{figure}[ht]
\centering
\providecommand{\GCOPDValidationFigureWidth}{0.8\linewidth}
\includegraphics[width=\GCOPDValidationFigureWidth]{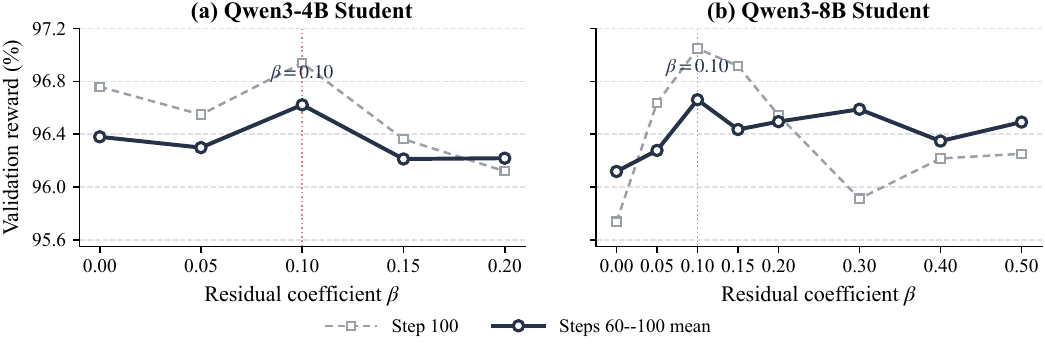}
\caption{Held-out validation used to select the residual coefficient $\beta$. The fixed set contains 231 GoLongRL examples after 32K-token filtering and has no overlap with training. White squares show the step-100 reward. Circles show the mean over steps 60, 70, 80, 90, and 100. Both summaries select $\beta=0.10$ at both model scales.}
\label{fig:beta-validation}
\end{figure}

\paragraph{Residual-coefficient selection.}
We select a single residual coefficient for both model scales using a fixed GoLongRL holdout. Before constructing the training subset, we reserve the first 256 examples in the ordered GoLongRL shards. Applying the same 32K-token limit leaves 231 validation examples with no overlap with the 9{,}527 training examples. Figure~\ref{fig:beta-validation} reports the reward at the final checkpoint and the mean over the last five validation checkpoints, using one stochastic response per example. Both summaries select $\beta=0.10$ for Qwen3-4B and Qwen3-8B, so we use this shared value in the main experiments. Because the holdout contains only the High-Recall Retrieval task family, we use it exclusively for coefficient selection rather than downstream evaluation.

\ifdefined\GCOPDSingleColumnLayout

\fi

\subsection{Main Results}

\paragraph{\method{} improves performance across model scales.}
As shown in Table~\ref{tab:main-results}, \method{} achieves the highest five-task average for both students. Relative to vanilla OPD, it raises the average score from 39.31 to 40.47 for Qwen3-4B and from 43.56 to 44.65 for Qwen3-8B. It also achieves the highest average among the evaluated shared-setup implementations at both model scales. Competing methods often lead on individual benchmarks but do not maintain the strongest aggregate performance. \method{} therefore provides a more favorable balance across the heterogeneous evaluation suite rather than relying on a single task.

\paragraph{Gains concentrate on structured reasoning and evidence aggregation.}
Relative to vanilla OPD, \method{} improves DocMath, MRCR, and CorpusQA for both students, with the largest gains on CorpusQA. These benchmarks emphasize structured reasoning and evidence aggregation. The concentration of gains is consistent with the intended role of response-level verifier calibration, although the table alone does not isolate task-specific mechanisms. Frames and LBv1QA show smaller or model-dependent changes, so the benefit is not uniform across tasks. Improvements over Raw confirm the effectiveness of the complete long-context post-training pipeline, whereas comparisons with vanilla OPD isolate the additional contribution of \method{}. The following ablations separately test the signed residual and its token-allocation rule.

\subsection{Ablation Studies}

We next examine the two components introduced by GC-OPD: the signed residual and RACA token allocation. All ablations use independently trained Qwen3-8B students under the shared setup and 100-step budget. We first vary the signal added to vanilla OPD while holding the coefficient and token-allocation rule fixed. We then examine how the residual should be distributed across tokens. Detailed formulations of all ablation variants are provided in Appendix~\ref{sec:ablation-configurations}.

\paragraph{Residualization improves upon OPD-only and direct-reward controls.}
Table~\ref{tab:residual-signal-ablation} isolates the signal added to vanilla OPD. The three augmented variants share RACA and $\beta=0.10$, so only the added term changes:
\begin{equation}
A'^{(i)}_t=A^{(i)}_t+
\begin{cases}
0, & \text{Vanilla OPD},\\
\beta c^{(i)}_t A^{(i)}_t, & \text{Additional OPD},\\
\beta c^{(i)}_t\tilde{R}^{(i)}, & \text{Direct reward},\\
\beta c^{(i)}_t\rho^{(i)}, & \text{GC-OPD}.
\end{cases}
\label{eq:signal-ablation-updates}
\end{equation}
Additional OPD changes the average score only from 43.56 to 43.60, indicating that another OPD-derived term does not explain the improvement. Direct reward reaches 44.19, confirming that group-normalized verifier feedback complements dense OPD guidance. Replacing the direct reward with the signed residual further raises the average score to 44.65 under the same coefficient and token-allocation rule. This additional gain supports accounting for the relative preference already expressed by OPD rather than adding verifier feedback alone.

\ifdefined\GCOPDSingleColumnLayout
\begin{table}[ht]
\else
\begin{table*}[ht]
\fi
\caption{Response-level signal ablation on Qwen3-8B. All augmented variants use RACA with coefficient $\beta=0.10$ and differ only in whether the added signal is $A^{(i)}_t$, $\tilde{R}^{(i)}$, or $\rho^{(i)}=\tilde{R}^{(i)}-\tilde{s}^{(i)}$. ``Avg.'' is the unweighted five-task mean, and ``$\Delta$'' is computed from the unrounded average relative to vanilla OPD.}
\label{tab:residual-signal-ablation}
\centering
\small
\setlength{\tabcolsep}{4.0pt}
\begin{tabular}{l c c c c c c c}
\toprule
\bf Variant & \bf Avg. & \bf DocMath & \bf Frames & \bf MRCR & \bf CorpusQA & \bf LBv1QA & $\boldsymbol{\Delta}$ \\
\cmidrule(lr){1-1} \cmidrule(lr){2-8}
Vanilla OPD & 43.56 & 55.13 & 34.59 & 30.44 & 39.82 & 57.80 & -- \\
Additional OPD & 43.60 & 54.63 & 34.59 & 30.03 & 39.51 & 59.20 & $+0.04$ \\
Direct reward & 44.19 & 55.13 & 34.59 & 30.40 & 40.73 & \textbf{60.10} & $+0.63$ \\
\cmidrule(lr){1-1} \cmidrule(lr){2-8}
\textbf{GC-OPD} & \textbf{44.65} & \textbf{55.50} & 34.59 & \textbf{31.10} & \textbf{43.77} & 58.30 & $\mathbf{+1.10}$ \\
\bottomrule
\end{tabular}
\ifdefined\GCOPDSingleColumnLayout
\end{table}
\else
\end{table*}
\fi

\paragraph{RACA improves residual allocation.}
Having established the contribution of the residual, we next examine its token allocation. Table~\ref{tab:token-credit-ablation} uses vanilla OPD as the no-residual anchor. All residual variants use $A'^{(i)}_t=A^{(i)}_t+0.10c^{(i)}_t\rho^{(i)}$ and differ only in $c^{(i)}_t$. Uniform sets $c^{(i)}_t=1$. Absolute OPD assigns credit using
\begin{equation}
c^{(i)}_t=\operatorname{clip}_{[0,5]}\!\left(
\frac{\lvert A^{(i)}_t\rvert}
{(T^{(i)})^{-1}\sum_{v=1}^{T^{(i)}}\lvert A^{(i)}_v\rvert}
\right),
\label{eq:absolute-opd-allocation}
\end{equation}
whereas RACA uses Equation~\ref{eq:raca}.

Uniform allocation raises the average score from 43.56 to 44.28, showing that the signed residual is useful even without token-dependent credit. RACA further raises it to 44.65, whereas Absolute OPD reaches only 43.93. Because Absolute OPD discards the sign of the OPD advantage, it can assign large credits to strongly negative OPD values. This result supports preserving the signed within-response ordering used by RACA.

\ifdefined\GCOPDSingleColumnLayout
\begin{table}[ht]
\else
\begin{table*}[ht]
\fi
\caption{Token-credit ablation on Qwen3-8B. All residual variants use the same signed residual with $\beta=0.10$ and differ only in token allocation. ``Avg.'' is the unweighted five-task mean, and ``$\Delta$'' is computed from the unrounded average relative to vanilla OPD.}
\label{tab:token-credit-ablation}
\centering
\small
\setlength{\tabcolsep}{4.0pt}
\begin{tabular}{l c c c c c c c}
\toprule
\bf Token allocation & \bf Avg. & \bf DocMath & \bf Frames & \bf MRCR & \bf CorpusQA & \bf LBv1QA & $\boldsymbol{\Delta}$ \\
\cmidrule(lr){1-1} \cmidrule(lr){2-8}
Vanilla OPD & 43.56 & 55.13 & 34.59 & 30.44 & 39.82 & 57.80 & -- \\
Absolute OPD & 43.93 & 54.63 & 35.56 & 29.53 & 41.34 & 58.60 & $+0.38$ \\
Uniform & 44.28 & 54.12 & \textbf{37.01} & 30.64 & 40.12 & \textbf{59.50} & $+0.72$ \\
\cmidrule(lr){1-1} \cmidrule(lr){2-8}
\textbf{RACA} & \textbf{44.65} & \textbf{55.50} & 34.59 & \textbf{31.10} & \textbf{43.77} & 58.30 & $\mathbf{+1.10}$ \\
\bottomrule
\end{tabular}
\ifdefined\GCOPDSingleColumnLayout
\end{table}
\else
\end{table*}
\fi

\ifdefined\GCOPDSingleColumnLayout
\FloatBarrier
\fi

\subsection{Disagreement Across Task Families}

Beyond downstream performance, we use a fixed response set to test whether teacher--verifier disagreement extends beyond the two tasks in Figure~\ref{fig:teacher-verifier-motivation}. This diagnostic measures disagreement prevalence rather than trained-checkpoint performance.

\paragraph{Teacher--verifier disagreement extends across task families.}
Figure~\ref{fig:task-conditioned-mismatch} in Appendix~\ref{sec:task-conditioned-disagreement} reports pairwise disagreement and top-1 mismatch for four task families containing at least 500 prompts, together with a prompt-macro aggregate over all nine families. The Multi-Table Extraction and High-Recall Retrieval rows aggregate the same frozen responses used in Figure~\ref{fig:teacher-verifier-motivation} over prompts shorter than 32K. The additional task rows show that disagreement extends beyond these two tasks, while the variation across rows indicates that its prevalence remains task dependent.

\section{Conclusion}

We introduced \method{} to address teacher--verifier disagreement in long-context OPD. \method{} forms a signed residual between group-normalized verifier rewards and trajectory-level OPD scores, then uses RACA to distribute it across tokens. This preserves dense OPD guidance while supporting binary and graded rewards without cross-task calibration.

Diagnostics reveal stronger disagreement over longer inputs in two tasks and task-dependent prevalence across the training mixture. Across five benchmarks, \method{} improves vanilla OPD from 39.31 to 40.47 for Qwen3-4B and from 43.56 to 44.65 for Qwen3-8B. Ablations identify residual calibration as the main contributor, with an additional gain from RACA over uniform allocation. Overall, \method{} integrates response-level verification with dense token-level guidance.

\bibliographystyle{plainnat}
\bibliography{references_all}

\clearpage
\appendix
\section{Task-Conditioned Disagreement Analysis}
\label{sec:task-conditioned-disagreement}

The main paper analyzes how teacher--verifier disagreement changes with prompt
length in two representative tasks. Here, we complement that analysis with a
task-conditioned view of four GoLongRL task families containing at least 500
prompts. For Multi-Table Extraction and High-Recall Retrieval, we aggregate the
same frozen responses used in Figure~1 over prompts shorter than 32K, without
separating the two length ranges. Each of the other task rows is likewise
computed from eight fixed Qwen3-8B responses per prompt. Overall pools all nine
task families by giving each informative prompt equal weight.

Within each informative rollout group, the pairwise disagreement rate is the
fraction of response pairs with distinct verifier rewards for which the OPD and
verifier orderings disagree. We average this rate across prompts. The top-1
mismatch rate is the fraction of these groups in which the response with the
highest trajectory-level OPD score does not attain the group's highest verifier
reward.
Their variation across task families complements the length-conditioned trends
in the main paper and motivates comparing the two assessments locally within
each rollout group. Because the responses are fixed, the analysis characterizes
signal disagreement rather than task-specific training gains or causal effects
of task type or context length.

\begin{figure}[H]
\centering
\includegraphics[width=0.74\columnwidth]{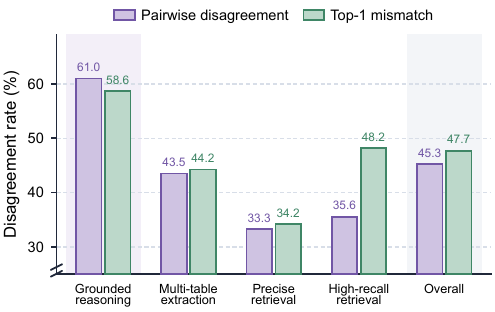}
\caption{Task-conditioned teacher--verifier disagreement in four GoLongRL task
families. Bars report prompt-macro pairwise disagreement rates and top-1
mismatch rates over rollout groups with at least two distinct verifier rewards.
Overall pools all nine task families.}
\label{fig:task-conditioned-mismatch}
\end{figure}

\FloatBarrier
\section{GC-OPD Implementation Details}

Using the notation of the main paper, $A^{(i)}_t$ is the detached token-level OPD
advantage, $s^{(i)}=(T^{(i)})^{-1}\sum_t A^{(i)}_t$ is its trajectory-level OPD score, and
$R^{(i)}$ is the verifier reward. The main paper defines the compact
pre-clipping update $A'^{(i)}_t$. Here, $T^{(i)}$ is the number of valid tokens
in response $i$, $G$ is the rollout-group size, and $\beta$ is the residual
coefficient in the main update. This section specifies the numerical guards and
update order used in the completed runs. The thresholds $\tau_G$ and $\tau_T$
guard group-level and token-level standard deviations, respectively, and
$a_{\max}$ is the final advantage-clipping bound. For $h\in\{R,s\}$, let
$\mu_h$ and $\sigma_h$ denote the population mean and standard deviation over
the $G$ responses. The implemented residual is
\begin{equation}
\rho^{(i)}=
\begin{cases}
\frac{R^{(i)}-\mu_R}{\sigma_R}
-\frac{s^{(i)}-\mu_s}{\sigma_s},
&\sigma_R,\sigma_s>\tau_G,\\
0,&\text{otherwise}.
\end{cases}
\label{eq:guarded-residual}
\end{equation}
If either group-level signal has negligible variation, the residual is zero
and the update reduces to vanilla OPD.

The token credit uses an analogous guard. Within response $i$, let
$\sigma_A^{(i)}$ denote the population standard deviation of its valid token
advantages. The guarded RACA credit is
\begin{align}
u^{(i)}_t&=
\begin{cases}
\dfrac{A^{(i)}_t-s^{(i)}}{\sigma_A^{(i)}},
&T^{(i)}\geq2\ \text{and}\ \sigma_A^{(i)}>\tau_T,\\
0,&\text{otherwise},
\end{cases}
\label{eq:normalized-token-advantage}\\
c^{(i)}_t&=1+\tanh\!\left(\frac{u^{(i)}_t}{2}\right),
\label{eq:implemented-token-credit}
\end{align}
Here, $u^{(i)}_t$ is the token's OPD advantage relative to the response mean,
measured in units of the within-response standard deviation. It describes
relative OPD support rather than token correctness. Thus negligible token
variation gives unit credit. After forming
$A'^{(i)}_t$, we clip it to $[-a_{\max},a_{\max}]$ before the policy update.
Table~\ref{tab:training} reports $\tau_G$, $\tau_T$, and $a_{\max}$. These
Given the OPD advantages and verifier rewards, these operations require no
additional teacher or student forward pass.

\begin{algorithm}[H]
\caption{Group-Calibrated On-Policy Distillation}
\label{alg:gcopd}
\begin{algorithmic}[1]
\REQUIRE Prompt batch $\mathcal{B}$; trainable student policy $\pi_\theta$; teacher $\pi_T$;
verifier $V$; group size $G$; residual coefficient $\beta$; thresholds
$\tau_G,\tau_T$; clipping bound $a_{\max}$
\ENSURE Updated student policy $\pi_\theta$
\STATE Set $\theta_{\mathrm{old}}\leftarrow\theta$ and freeze
$\pi_{\theta_{\mathrm{old}}}$ for rollout generation
\FOR{each prompt $\mathbf{x}\in\mathcal{B}$}
  \STATE Sample $\mathcal{G}(\mathbf{x})=\{\mathbf{y}^{(i)}\}_{i=1}^{G}$
  from $\pi_{\theta_{\mathrm{old}}}(\cdot\mid\mathbf{x})$
  \FOR{$i=1,\ldots,G$}
    \STATE $R^{(i)}\leftarrow V(\mathbf{x},\mathbf{y}^{(i)})$
    \STATE Compute detached $A^{(i)}_t$ under $\pi_T$ and
    $\pi_{\theta_{\mathrm{old}}}$ for
    $t=1,\ldots,T^{(i)}$
    \STATE $s^{(i)}\leftarrow (T^{(i)})^{-1}
    \sum_{t=1}^{T^{(i)}} A^{(i)}_t$
  \ENDFOR
  \STATE Compute population statistics $(\mu_R,\sigma_R)$ and
  $(\mu_s,\sigma_s)$ over the $G$ responses
  \IF{$\sigma_R>\tau_G \land \sigma_s>\tau_G$}
    \STATE $\rho^{(i)}\leftarrow
    \dfrac{R^{(i)}-\mu_R}{\sigma_R}
    -\dfrac{s^{(i)}-\mu_s}{\sigma_s}$ for $i=1,\ldots,G$
  \ELSE
    \STATE $\rho^{(i)}\leftarrow0$ for all $i$
  \ENDIF
  \FOR{each response $i$}
    \STATE Compute $\sigma_A^{(i)}$ from
    $\{A^{(i)}_t\}_{t=1}^{T^{(i)}}$
    \IF{$T^{(i)}\geq2 \land \sigma_A^{(i)}>\tau_T$}
      \STATE $u^{(i)}_t\leftarrow
      \dfrac{A^{(i)}_t-s^{(i)}}{\sigma_A^{(i)}}$
      for $t=1,\ldots,T^{(i)}$
      \STATE $c^{(i)}_t\leftarrow
      1+\tanh\!\left(u^{(i)}_t/2\right)$
      for $t=1,\ldots,T^{(i)}$
    \ELSE
      \STATE $u^{(i)}_t\leftarrow0$ and $c^{(i)}_t\leftarrow1$ for
      $t=1,\ldots,T^{(i)}$
    \ENDIF
    \STATE $A'^{(i)}_t\leftarrow A^{(i)}_t+
    \beta c^{(i)}_t\rho^{(i)}$ for $t=1,\ldots,T^{(i)}$
    \STATE $\widehat A^{(i)}_t\leftarrow
    \operatorname{clip}(A'^{(i)}_t,-a_{\max},a_{\max})$ for all $t$
  \ENDFOR
\ENDFOR
\STATE Update $\pi_\theta$ using the same policy surrogate as vanilla OPD, with
$\widehat A^{(i)}_t$ as the token-level advantages
\end{algorithmic}
\end{algorithm}

\section{Experimental Details}

\subsection{Models, Data, and Training Configuration}

We initialize Qwen3-4B and Qwen3-8B students from their official checkpoints
and use Qwen3-30B-A3B-Thinking-2507 as the teacher. Students generate in
no-thinking mode, and the teacher scores the sampled response tokens without
generating separate responses.

The 32K-filtered GoLongRL training file contains 9,527 prompts.
Table~\ref{tab:data} reports its prompt-length distribution, complementing the
task-family and verifier statistics in the main paper. The median, 90th
percentile, and maximum lengths are 9,923, 26,940, and 32,766 tokens.

\begin{table}[t]
\captionsetup{justification=raggedright,singlelinecheck=false,skip=6pt}
\caption{Prompt-length distribution of the exact 32K-filtered training file.}
\label{tab:data}
\centering
{\small
\setlength{\tabcolsep}{6pt}
\renewcommand{\arraystretch}{1.05}
\begin{tabular*}{0.36\textwidth}{@{\extracolsep{\fill}}cc@{}}
\toprule
\textbf{Prompt length} & \textbf{Samples} \\
\midrule
$\leq 2$K & 2,222 \\
2--8K & 1,958 \\
8--16K & 2,155 \\
16--24K & 1,914 \\
24--32K & 1,278 \\
\midrule
Total & 9,527 \\
\bottomrule
\end{tabular*}
}
\end{table}

Table~\ref{tab:training} records the common configuration. Raw rows receive no
training; every trained method uses this configuration and is evaluated at
its final checkpoint.

\begin{table}[t]
\captionsetup{justification=raggedright,singlelinecheck=false,skip=6pt}
\caption{Shared training configuration and GC-OPD settings.}
\label{tab:training}
\centering
{\small
\setlength{\tabcolsep}{4.0pt}
\renewcommand{\arraystretch}{1.05}
\ifdefined\GCOPDSingleColumnLayout
\begin{tabular}{@{}p{0.38\textwidth}>{\centering\arraybackslash}p{0.35\textwidth}@{}}
\else
\begin{tabular}{@{}p{0.45\columnwidth}>{\centering\arraybackslash}p{0.51\columnwidth}@{}}
\fi
\toprule
\textbf{Setting} & \textbf{Value} \\
\midrule
Training horizon & 100 steps \\
Prompt batch size & 32 \\
Responses per prompt & 8 \\
Maximum prompt length & 32,768 \\
Training response cap & 10,240 \\
Learning rate & $10^{-6}$ \\
Warmup & 0 \\
Weight decay & 0.01 \\
Policy mini-batch size & 4 \\
PPO epochs & 1 \\
PPO clip ratio & 0.2 \\
Loss aggregation & Token mean \\
Rollout temperature & 1 \\
Rollout top-$p$ & 1 \\
Precision & bfloat16 \\
Random seed & 42 \\
Residual coefficient & $\beta=0.10$ \\
Group std. threshold & $\tau_G=10^{-6}$ \\
Token std. threshold & $\tau_T=10^{-6}$ \\
Final advantage clip & $[-10,10]$ ($a_{\max}=10$) \\
\bottomrule
\end{tabular}
}
\end{table}

\paragraph{Compute resources.}
Each training run used eight 80-GB NVIDIA H800 or H100 GPUs. Rollout tensor
parallelism and actor/reference sequence parallelism were both set to 8.

\subsection{Ablation Configurations}
\label{sec:ablation-configurations}

The ablations in the main paper use independent Qwen3-8B training runs under
the shared configuration in Table~\ref{tab:training}. All runs use the same
student initialization, teacher, filtered training data, optimizer, rollout
budget, and 100-step training horizon. They are evaluated at step 100 with the
same five-benchmark pipeline as the main comparison.

\paragraph{Added-signal ablation.}
Vanilla OPD provides the common anchor for this comparison. The three augmented
variants use the same RACA credit and a scalar coefficient of 0.10, while
changing the signal added to the vanilla OPD advantage. Their updates before
final advantage clipping are
\begin{align}
A'^{(i)}_t&=A^{(i)}_t
&&\text{(vanilla OPD)},\\
A'^{(i)}_t&=A^{(i)}_t+\beta c^{(i)}_tA^{(i)}_t,
\quad \beta=0.10
&&\text{(Additional OPD)},\\
A'^{(i)}_t&=A^{(i)}_t+\beta c^{(i)}_t\tilde{R}^{(i)},
\quad \beta=0.10
&&\text{(Direct reward)},\\
A'^{(i)}_t&=A^{(i)}_t+\beta c^{(i)}_t\rho^{(i)},
\quad \beta=0.10
&&\text{(GC-OPD)}.
\end{align}
The comparison holds the original OPD signal, RACA allocation, and coefficient
fixed. Additional OPD adds another credit-modulated OPD term, Direct reward uses the group-normalized verifier reward
$\tilde{R}^{(i)}$, whereas GC-OPD subtracts the group-normalized trajectory
OPD score and uses $\rho^{(i)}=\tilde{R}^{(i)}-\tilde{s}^{(i)}$. It
therefore separates the effect of adding another OPD-derived term from adding verifier feedback, and tests whether accounting for the relative preference already expressed by OPD improves upon adding verifier feedback directly.

\paragraph{Token-allocation ablation.}
This comparison fixes the signed residual and the same $\beta=0.10$ used by
the main configuration, so every calibrated variant uses
$A'^{(i)}_t=A^{(i)}_t+0.10c^{(i)}_t\rho^{(i)}$. Only the token credit
$c^{(i)}_t$ changes. Uniform allocation sets
$c^{(i)}_t=1$. Absolute OPD is a token-allocation control, not an alternative
trajectory-level signal. It assigns credit in proportion to the magnitude of
the tokenwise OPD advantage:
\begin{equation}
c^{(i)}_t=\operatorname{clip}_{[0,5]}\!\left(
\frac{\lvert A^{(i)}_t\rvert}
{(T^{(i)})^{-1}\sum_{v=1}^{T^{(i)}}\lvert A^{(i)}_v\rvert}
\right),
\label{eq:absolute-opd-credit}
\end{equation}
The upper bound of 5 limits extreme token weights, and unit credit is used when
the denominator is numerically negligible. RACA uses
Equation~\ref{eq:implemented-token-credit}. Thus this comparison holds the
trajectory-level residual and its coefficient fixed while isolating how the
correction is distributed across response tokens.

\subsection{Baselines under the Shared Setup}

All trained baselines share the student initialization, teacher, filtered
training data, rollout budget, optimizer, training horizon, and clipped
surrogate. We hold these factors fixed to isolate each method's principal
training-signal mechanism. The comparisons should therefore be interpreted as
controlled mechanism comparisons under a shared setup rather than exact
reproductions of the original experimental recipes. Each method generates its
own trajectories on-policy. Let
$p_{T,t}$, $p_{S,t}$, and $p_{0,t}$ denote the teacher, current-student, and
frozen student-base probabilities of the sampled token. The paragraphs below
record only method-specific implementation choices. To avoid clutter, the
response index $i$ is suppressed in baseline-specific token formulas unless
it is needed explicitly.

\paragraph{Raw checkpoint.}
Raw evaluates the official Qwen3-4B or Qwen3-8B checkpoint without additional
training, using the same downstream evaluation pipeline.

\paragraph{OPD.}
OPD uses $A_t^{\mathrm{OPD}}=\log p_{T,t}-\log p_{S,t}$ as the detached token
advantage. The native verifier is executed by the shared pipeline, but its
terminal score is not added to this advantage. Equivalently, vanilla OPD is
recovered from GC-OPD by setting $\beta=0$, and we use this convention
throughout the supplement.

\paragraph{ExOPD.}
ExOPD replaces the teacher log-probability target with an extrapolated target
anchored at the corresponding frozen student-base checkpoint:
\begin{equation}
A_t^{\mathrm{Ex}}=
\lambda\log p_{T,t}+(1-\lambda)\log p_{0,t}-\log p_{S,t}.
\label{eq:exopd}
\end{equation}
Our runs use $\lambda=1.25$ and the corresponding frozen Qwen3 initialization
as $p_0$, following the default student-base-reference formulation. The
reference contributes only to the detached token target and is not used as a
KL regularizer. In the ExOPD paper, ``reward correction'' denotes the optional
replacement of $p_0$ with the teacher's pre-RL base model. This reference-based
variant is distinct from the verifier reward used by GC-OPD and is not used
here.

\paragraph{Uni-OPD.}
Uni-OPD combines token-level teacher supervision with response-level outcome
separation. Within the eight responses for one prompt, responses with positive
verifier reward form the correct set and the others form the
incorrect set. Let $\bar{s}^{+}$ and $\bar{s}^{-}$ be the mean response-level
OPD scores of these two sets. If both sets exist, the required shift is
\begin{equation}
d=\max\{0,\,\delta-(\bar{s}^{+}-\bar{s}^{-})\}.
\label{eq:uni-margin}
\end{equation}
With bidirectional calibration, $d/2$ is added to every token of a correct
response and subtracted from every token of an incorrect response. Groups
lacking either set receive no margin shift.

For a controlled comparison, we isolate Uni-OPD's outcome-guided margin
calibration and omit its offline and online data-balancing components. This
keeps the training data and rollout budget consistent across methods. The
implementation uses group scope, mean statistics, $\delta=0.4$, and
bidirectional shifts. Following the released implementation, the resulting
group-level shift is added to the original tokenwise OPD advantages. We also
retain the released configuration's teacher--student log-probability-gap mask
with a threshold of 10 as a training-stability setting.

\paragraph{PowerOPD.}
PowerOPD changes the geometry of the token signal from a log-probability gap to
a bounded probability-power difference:
\begin{equation}
A_t^{\mathrm{Power}}=
\operatorname{sg}\!\left[p_{T,t}^{\alpha}-p_{S,t}^{\alpha}\right],
\qquad \alpha=100,
\label{eq:poweropd}
\end{equation}
where $\operatorname{sg}$ denotes stop-gradient. We use this bounded signal as
the method-specific token advantage within the shared clipped surrogate. Both
model scales use $\alpha=100$.

\paragraph{FiRe-OPD.}
Following the released implementation, FiRe-OPD computes the length-normalized
teacher score $q^{(i)}=(T^{(i)})^{-1}\sum_t\log p_{T,t}^{(i)}$ for each
trajectory and applies the actor-micro-batch percentile filter. Token weights
use teacher confidence and student confusion,
\begin{equation}
c_{T,t}=1-\frac{H_{T,t}}{\max H_T},\qquad
c_{S,t}=\frac{H_{S,t}}{\max H_S},
\end{equation}
clipped to $[0,1]$. The OPD advantage is multiplied by
$w_t=(1+\alpha_F c_{T,t})(1+\beta_F c_{S,t})$ after normalizing $w$ to mean one
over valid tokens. Filtered trajectories contribute no policy loss.

Both runs use a 20th-percentile trajectory cutoff and
$\alpha_F=\beta_F=1$. The percentile, entropy maxima, and weight normalization
are recomputed within each actor micro-batch supplied to the loss. The actor
starts from four-response policy mini-batches and may split them further under
dynamic token budgeting.

\subsection{Evaluation Implementation}

Table~\ref{tab:evaluation} records the evaluation units and aggregation rules.
The paragraphs below specify the per-example scorers.

\begin{center}
\begin{minipage}{\columnwidth}
\centering
\captionof{table}{Evaluation units and aggregation for the five reported
long-context benchmark columns.}
\label{tab:evaluation}
{\small
\setlength{\tabcolsep}{2.5pt}
\begin{tabular}{>{\raggedright\arraybackslash}p{0.18\columnwidth}
                >{\raggedright\arraybackslash}p{0.34\columnwidth}
                >{\raggedright\arraybackslash}p{0.38\columnwidth}}
\toprule
Benchmark & Evaluation units & Reported aggregation \\
\midrule
DocMath & Four splits: 200/100/200/300
        & Per-item max(rule, judge); four-split aggregate \\
FRAMES & 824 questions
       & Mean max(CEM, judge) accuracy \\
MRCR & 0--128K $\times$ \{2, 4, 8\} needles
     & Mean prefix-gated sequence ratio \\
CorpusQA & 0--128K $\times$ four domains
         & Overall example accuracy \\
LBv1QA & Five subsets $\times$ 200
       & Macro mean of five subset accuracies \\
\bottomrule
\end{tabular}
}
\end{minipage}
\end{center}

\paragraph{DocMath.}
The evaluator extracts the final answer and applies both the rule-based
numerical/equation checker and a semantic-equivalence judge. The per-example
score is the maximum of these binary decisions. Scores are then aggregated
over the four simple/complex and short/long splits shown in
Table~\ref{tab:evaluation}.

\paragraph{FRAMES.}
After removing any hidden reasoning segment, the scorer applies cover exact
match (CEM). CEM lowercases text, removes punctuation and articles, normalizes
comma-separated numbers, and requires the target as a word-bounded span in the
answer. A sample is correct if either CEM or the semantic-equivalence judge
accepts it.

\paragraph{MRCR.}
Every gold response begins with a sample-specific random prefix. A prediction
without the exact prefix receives zero. Otherwise, the scorer removes the
prefix and computes Python character-level \texttt{SequenceMatcher} similarity
between the prediction and reference. The aggregate includes the 2-, 4-, and
8-needle variants across context lengths.

\paragraph{CorpusQA.}
We use the 0--128K file covering Chinese finance, English finance, English
education, and English real estate. The evaluator extracts text following the
prescribed answer marker when present, then obtains a binary
semantic-equivalence judgment against the generated reference. The reported
column is overall example accuracy across the four domains.

\paragraph{LongBench v1 QA.}
The evaluator extracts the final occurrence of ``Therefore, the answer is''
and falls back to the last non-empty line when the marker is absent. It compares
the extracted answer with every accepted reference and takes the maximum binary
semantic judgment. The reported score is the macro mean over NarrativeQA,
Qasper, HotpotQA, 2WikiMultihopQA, and MuSiQue.

Student serving accepts at most 120,000 input tokens and uses a 131,072-token
model length with YaRN factor 4 and target length 131,072. The completed
evaluation runs cap generation at 8,192 tokens. Inputs that exceed the serving
budget are middle truncated so that both the beginning and end are retained.
Students remain in no-thinking mode, and no method receives a method-specific
evaluation prompt or decoding override. Semantic judging uses
Qwen3-30B-A3B-Instruct-2507 with a 32,768-token context and a 2,048-token
output cap. The reported aggregate is the naive average of the five benchmark
scores.

\section{Auditable Long-Context Case Study}

To make the GC-OPD signal computation concrete, we present one example from
GoLongRL. Its Qwen3-8B no-thinking prompt contains 17,265 tokens and asks which
matrix receives the largest relative improvement from parallel NUMA
optimization: A \texttt{rajat31}, B \texttt{HV15R}, C \texttt{cage15}, or D
\texttt{ldoor}. The dataset's native label is B. The contextual observations
reproduced in Figure~\ref{fig:case} support this label, but do not by themselves
constitute a direct speedup measurement.

This case is an illustrative audit of the signal computation rather than
additional quantitative evidence. We independently reroll the example using
the official Qwen3-8B base checkpoint and Qwen3-30B-A3B-Thinking-2507 as the
teacher. We sample eight responses with temperature 1, top-$p=1$, top-$k=-1$,
seed 42, and a 10,240-token response cap.
Table~\ref{tab:case-tokens} reports representative token signals recomputed
offline with $\beta=0.10$ and $a_{\max}=10$. In the table, the response index
is suppressed within each response block, and
$\Delta A_t=\beta c_t\rho$ denotes the additive residual correction before
clipping.

\ifdefined\GCOPDSingleColumnLayout
\begin{figure}[t]
\else
\begin{figure*}[!t]
\fi
\centering
\ifdefined\GCOPDSingleColumnLayout
\makebox[\textwidth][c]{%
\includegraphics[width=1.08\textwidth]{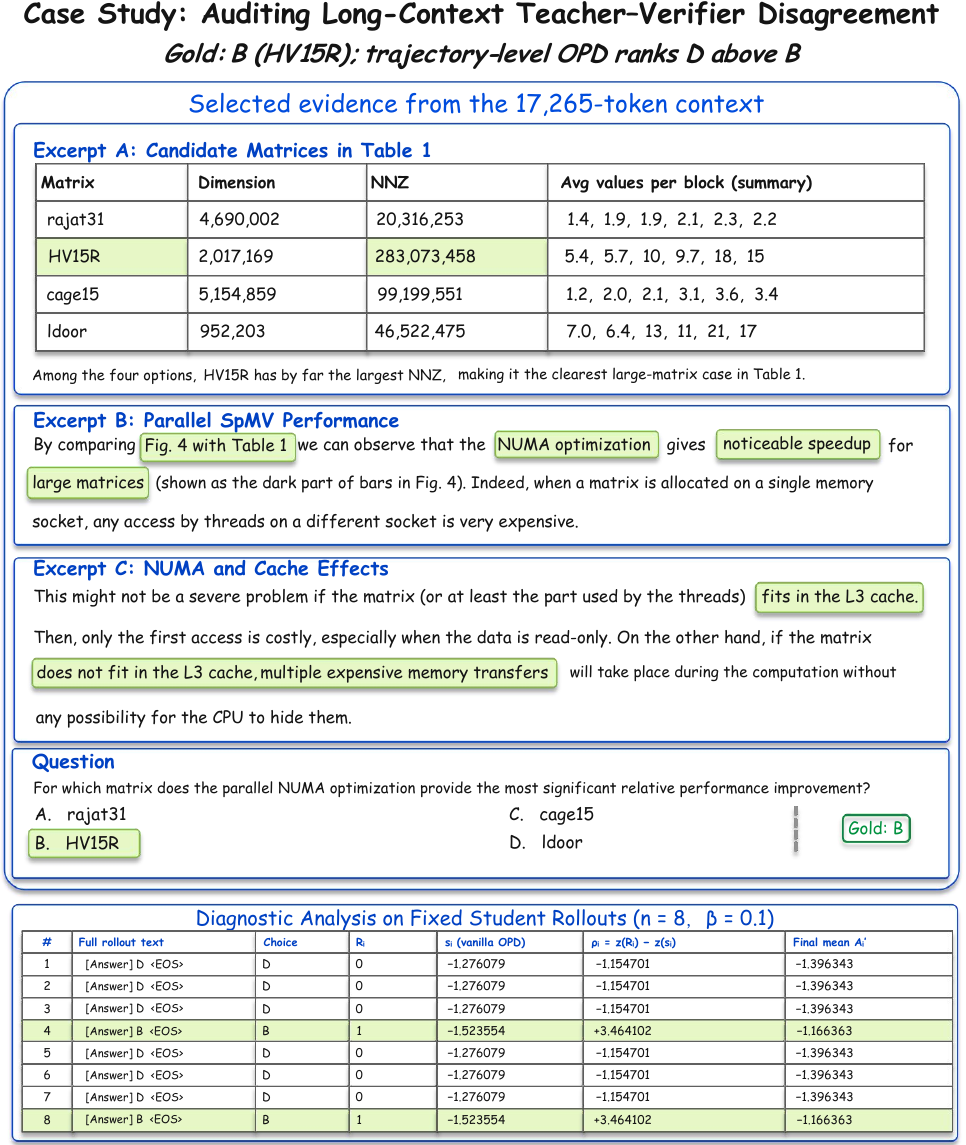}%
}
\else
\includegraphics[width=0.96\textwidth]{Figures/gcopd_case_study/gcopd_long_context_case_study.pdf}
\fi
\caption{Illustrative fixed-rollout visualization for the 17,265-token case.
It reproduces the selected context evidence and all eight independently
rerolled student responses; the native label is B.}
\label{fig:case}
\ifdefined\GCOPDSingleColumnLayout
\end{figure}
\else
\end{figure*}
\fi

\ifdefined\GCOPDSingleColumnLayout
\begin{table}[H]
\else
\begin{table*}[!t]
\fi
\caption{Representative token-level RACA recalculations for the correct-B and
incorrect-D responses in Figure~\ref{fig:case} under the final GC-OPD setting.
Values are rounded; ranking checks use unrounded values.}
\label{tab:case-tokens}
\centering
{\small
\setlength{\tabcolsep}{2.15pt}
\resizebox{0.96\textwidth}{!}{%
\begin{tabular}{@{}lrrrrr@{\hspace{18pt}}lrrrrr@{}}
\toprule
\multicolumn{6}{c}{\textbf{Correct B response}, $\rho_{\mathrm{B}}=+3.464102$} &
\multicolumn{6}{c}{\textbf{Incorrect D response}, $\rho_{\mathrm{D}}=-1.154701$} \\
\cmidrule(lr){1-6}\cmidrule(lr){7-12}
Token & $A_t$ & $u_t$ & $c_t$ & $\Delta A_t$ & $\widehat A_t$ &
Token & $A_t$ & $u_t$ & $c_t$ & $\Delta A_t$ & $\widehat A_t$ \\
\midrule
\texttt{[} & -4.88599 & -1.85923 & 0.26959 & 0.09339 & -4.79260 &
\texttt{[} & -4.88599 & -1.98073 & 0.24248 & -0.02800 & -4.91399 \\
\texttt{Answer} & -0.02188 & 0.83034 & 1.39285 & 0.48250 & 0.46062 &
\texttt{Answer} & -0.02188 & 0.68817 & 1.33112 & -0.15370 & -0.17558 \\
\texttt{]} & -0.64095 & 0.48803 & 1.23928 & 0.42930 & -0.21165 &
\texttt{]} & -0.64095 & 0.34849 & 1.17250 & -0.13539 & -0.77634 \\
\texttt{B} & -1.91057 & -0.21400 & 0.89341 & 0.30949 & -1.60108 &
\texttt{D} & -0.66057 & 0.33773 & 1.16728 & -0.13479 & -0.79535 \\
\texttt{EOS} & -0.15839 & 0.75486 & 1.36047 & 0.47128 & 0.31290 &
\texttt{EOS} & -0.17101 & 0.60634 & 1.29421 & -0.14944 & -0.32046 \\
\bottomrule
\end{tabular}%
}
}
\ifdefined\GCOPDSingleColumnLayout
\end{table}
\else
\end{table*}
\fi

\end{document}